\documentclass[letterpaper]{article} 
\usepackage{aaai25}  
\usepackage{times}  
\usepackage{helvet}  
\usepackage{courier}  
\usepackage[hyphens]{url}  
\usepackage{graphicx} 
\usepackage{natbib}  
\usepackage{caption} 
\usepackage{algorithm}
\usepackage{algorithmic}
\usepackage{amsmath}
\usepackage{xcolor}
\usepackage{amsfonts}
\usepackage{pifont}
\usepackage{multirow}
\usepackage{amsmath}
\usepackage{bm}
\usepackage{tikz}
\usepackage{newfloat}
\usepackage{listings}
\usepackage{array}
\usepackage[pagewise]{lineno}
\usepackage{subcaption}
\usepackage{amssymb}
\usepackage{multirow}
\usepackage{booktabs}
\usepackage{colortbl} 
\usepackage{enumitem}
\usepackage{array}
\usepackage{colortbl}
\usepackage{xcolor}
\definecolor{tan}{rgb}{0.937, 0.902, 0.843}
\usepackage{arydshln}
\usepackage{natbib}
\usepackage{amsthm}

\newtheorem{theorem}{Theorem}

\usepackage{newfloat}
\usepackage{listings}
\DeclareCaptionStyle{ruled}{labelfont=normalfont,labelsep=colon,strut=off} 
\floatstyle{ruled}
\newfloat{listing}{tb}{lst}{}
\floatname{listing}{Listing}
\title{Smoothness Really Matters: A Simple Yet Effective Approach for \\ Unsupervised Graph Domain Adaptation}
\author {
	Wei Chen\textsuperscript{\rm1},
		Guo Ye\textsuperscript{\rm2},
			Yakun Wang\textsuperscript{\rm2},
	Zhao Zhang\textsuperscript{\rm3},
		Libang Zhang\textsuperscript{\rm2}, \\
				Daixin Wang\textsuperscript{\rm2},
								Zhiqiang Zhang\textsuperscript{\rm2},
	Fuzhen Zhuang\textsuperscript{\rm1,4\thanks{Corresponding Author.}},
}
\affiliations {
	\textsuperscript{\rm 1}Institute of Artificial Intelligence, Beihang University, Beijing, China\\
	\textsuperscript{\rm 2}Independent Researcher, Beijing, China\\
	\textsuperscript{\rm 3}Institute of Computing Technology, Chinese Academy of Sciences, Beijing, China\\
	\textsuperscript{\rm 4}Zhongguancun Laboratory, Beijing, China\\
	\{chenwei23,zhuangfuzhen\}@buaa.edu.cn, zhangzhao2021@ict.ac.cn
}

\usepackage{bibentry}

\begin{document}

\maketitle

\begin{abstract}
	Unsupervised Graph Domain Adaptation (UGDA) seeks to bridge distribution shifts between domains by transferring knowledge from labeled source graphs to given unlabeled target graphs. Existing UGDA methods primarily focus on aligning features in the latent space learned by graph neural networks (GNNs) across domains, often overlooking structural shifts, resulting in limited effectiveness when addressing structurally complex transfer scenarios.
	Given the sensitivity of GNNs to local structural features, even slight discrepancies between source and target graphs could lead to significant shifts in node embeddings, thereby reducing the effectiveness of knowledge transfer.
	To address this issue, we introduce a novel approach for UGDA called Target-Domain Structural Smoothing (TDSS). 
	 TDSS is a simple and effective method designed to perform structural smoothing directly on the target graph, thereby mitigating structural distribution shifts and ensuring the consistency of node representations. 
	 Specifically, by integrating smoothing techniques with neighborhood sampling, TDSS maintains the structural coherence of the target graph while mitigating the risk of over-smoothing.
 Our theoretical analysis shows that TDSS effectively reduces target risk by improving model smoothness. Empirical results on three real-world datasets demonstrate that TDSS outperforms recent state-of-the-art baselines, achieving significant improvements across six transfer scenarios. The code is available in https://github.com/cwei01/TDSS.

\end{abstract}

\section{Introduction}
Graph Neural Networks (GNNs) have become a powerful tool for processing graph-structured data~\cite{zhu2023wingnn,wang2024optimizing,zhu2024incorporating,liao2024revgnn,chen2024fairgap,chen2024fairdgcl,liang2024survey,liang2024mines}.
Despite their notable success, GNNs often struggle with generalizing to distribution shifts, leading to unsatisfactory performance in the new domains~\cite{li2022ood,gao2023survey,luo2024ci4rs}.

To tackle the challenge of distribution shifts across domains, Unsupervised Graph Domain Adaptation (UGDA)~\cite{wilson2020survey,shi2024graph} provides an effective strategy by focusing on minimizing domain discrepancies, enabling the transfer of knowledge from a well-labeled source domain to a target domain abundant in unlabeled samples.
Inspired by domain adaptation success in computer vision (CV)~\cite{xu2024beyond} and natural language processing (NLP)~\cite{xu2023retrieval}, most UGDA methods use GNNs to create dense node representations and apply regularization techniques like Maximum Mean Discrepancy (MMD)~\cite{shen2020network,liu2024pairwise}, graph subtree discrepancy \cite{wu2023non}, and adversarial learning \cite{xiao2023spa,li2024comprehensive} to maintain consistency across source and target domains.

\begin{figure}[t]
	\vspace{-0pt}
	\centering
	\setlength{\fboxrule}{0.pt}
	\setlength{\fboxsep}{0.pt}
	\fbox{
		\includegraphics[width=0.95\linewidth]{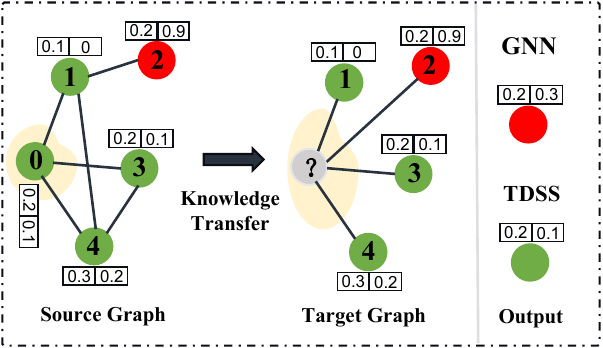}
	}
	\caption{An example of GNNs sensitivity to similar inputs. Nodes have two features; green and red represent different classes. We assume the second feature determines node class: values $\leq 0.2$ are green, and values $> 0.2$ are red.
	}
	\label{intro}
	\vspace{-5pt}
\end{figure}

Nonetheless, these approaches often overlook the distinct impacts of distribution shifts driven by graph structures, which can yield suboptimal results. 
For instance, when performing domain adaptation between the \textit{Cora} and \textit{PubMed} citation networks\footnote{They are widely used citation networks: \textit{Cora} focuses on machine learning papers, while \textit{PubMed} covers biomedical research.}~\cite{chen2010community}, it's crucial to consider the structural differences. Network motifs \cite{alon2007network}, which are specific subgraph patterns capturing local characteristics, offer a means to quantify these differences. Specifically,
in \textit{Cora}, about 30\% of the structures are triangular motifs (i.e., \scalebox{1}{$ \boldsymbol \triangle$}), 
indicating tight citation clusters and suggesting knowledge propagation within small, cohesive communities.
In contrast, the \textit{PubMed} features star-shaped motifs (i,e., \scalebox{1.5}{$\ast$}), where many papers cite a few core papers, indicating reliance on a few influential works with limited direct connections among the citing papers.
Although both networks involve overlapping research topics and citation patterns, 
the differences in their structural motifs could  impact the effectiveness of domain adaptation strategies.

Conversely, in image and text domain~\cite{chen2024modeling,xu2024end}, such similarities in feature distributions or content patterns can yield modest performance gains during knowledge transfer. This contrast underscores the fundamental differences in domain adaptation between graph structures and other data types. 
The reason lies in GNNs' pronounced sensitivity to local structural stability, where even slight discrepancies between source and target graphs can result in significant shifts in node embeddings, thereby diminishing the effectiveness of knowledge transfer.
Consider Figure~\ref{intro} as an example. In the source graph, the input neighbor vectors \([0.1, 0]\), \([0.2, 0.1]\), and  \([0.3, 0.2]\) result in a GNNs output vector of \([0.2, 0.1]\). However, in the target graph, despite a similar input, the GNNs produces a distinctly different output vector of \([0.2, 0.3]\) due to slight structural changes in neighbor node. 
This highlights the need for a method to constrain local variations and improve model robustness.
In light of this, a few recent studies~\cite{liu2023structural,liu2024pairwise} have primarily addressed structural shifts by reweighting edges to modify the source graph's structure, thereby aligning it more closely with target graph. However, it may fall short when significant differences in nodes and edges exist between domains, as it might not fully capture target graph's structural characteristics, particularly when transferring from a simpler graph to a more complex one.


Smoothing techniques~\cite{field1988laplacian}, known for their ability to reduce inter-node discrepancies and improve the robustness of GNNs models, have been successfully applied in various tasks like node classification and link prediction. Inspired by the aforementioned analysis, this paper proposes a simple yet effective approach for UGDA called Target-Domain Structural Smoothing (TDSS) to address structural shifts problem.
Our method is adaptable to various complex transfer scenarios. Specifically, by applying structural smoothing directly to the target graph, TDSS better accommodates the complex structures of the target domain. Specifically, it integrates a smoothing term with a neighborhood sampling strategy. 
The sampling mechanism flexibly explores the graph structure, avoiding over-smoothing while preserving key distinctions between nodes. Simultaneously, the smoothing term enhances model consistency, maintaining structural integrity in node representations and mitigating the effects of local perturbations and unstable features.
Our theoretical analysis shows that target risk is primarily influenced by both model smoothness and domain discrepancies.
Empirical results on three datasets demonstrate that TDSS outperforms recent state-of-the-art baselines.

Our major contributions are summarized as follows: \textbf{(Novel Perspective):} We are the first to approach UGDA through the lens of model smoothness. Our theoretical analysis reveals that target risk is predominantly driven by model smoothness and domain discrepancies, establishing a robust theoretical foundation for our method.  \textbf{(Simple yet Effective Method):}  We introduce TDSS, a straightforward and effective method that seamlessly integrates into existing frameworks, providing a plug-and-play solution to enhance model generalization. \textbf{(Experimental Results):} Our experiments demonstrate TDSS's adaptability and effectiveness across graph settings, surpassing state-of-the-art methods and consistently boosting performance in diverse transfer scenarios.

\section{Related Work}

\textbf{Unsupervised Graph Domain Adaptation.}
UGDA aims to transfer knowledge from a well-labeled source graph to an unlabeled target graph, tackling the distribution shifts inherent in graph-structured data. Unlike traditional domain adaptation techniques~\cite{zhuang2020comprehensive}, UGDA encounters greater challenges due to the non-IID nature of graph data, which complicates the alignment of node embeddings between the source and target graphs and can lead to mismatches and inconsistencies in the transferred knowledge. Consequently, many approaches have explored various techniques to achieve effective knowledge transfer, such as MMD \cite{shen2020network}, graph subtree discrepancy \cite{wu2023non}, and adversarial learning \cite{xiao2023spa,li2024comprehensive}. However, these methods often miss the specific challenges posed by structural shifts in graphs, which can result in less effective solutions.
Recently, a few studies have begun addressing distributional changes in graph structures by designing methods for structural alignment to reduce variations. For example, StruRW~\cite{liu2023structural} reweights edges in the source graph to mitigate neighborhood shifts, while PairAlign~\cite{liu2024pairwise} recalibrates node influence through edge weights and adjusts classification loss with label weights to manage both structural and label shifts.
Yet, when there is a significant disparity in the number of nodes and edges between the source and target graphs, direct edge reweighting may not fully capture the target graph's intricate structure and may still result in suboptimal performance.


\textbf{Model Smoothness.}
Model smoothness refers to designing a model that produces consistent and stable predictions across similar data points, ensuring that the prediction outputs are uniformly flat within certain neighborhoods~\cite{rosca2020case,wang2023direct}. This concept has already been both empirically and theoretically validated to improve model generalization and robustness in independent and identically distributed scenarios~\cite{wang2023direct}. In the UGDA setting, several approaches have sought to improve UGDA performance by incorporating additional model generalization constraints into their objectives. For example, SpecReg~\cite{you2023graph} derives a generalization bound for single-layer GNNs and enhances it by introducing spectral regularization. Similarly, A2GNN~\cite{liu2024rethinking} applies a Lipschitz formulation for $k$-layer GNNs, demonstrating that the target risk bound can be tightened by decreasing propagation layers in the source graph while increasing them in the target graph, though this approach can easily lead to over-smoothing.
Different from these approaches, our work focuses on directly reducing target risk by leveraging model smoothness, providing a novel perspective to enhance the effectiveness of UGDA.

\section{Problem Formulation}

Consider an undirected, unweighted graph \( \mathcal{G} = (\mathcal{V}, \mathcal{E}) \), where \( \mathcal{V} \) is the set of nodes connected by the edge set \( \mathcal{E} \). The adjacency matrix of \( \mathcal{G} \) is a binary symmetric matrix \(\mathbf{A}\) such that \(\mathbf{A}_{i,j} = 1\) if the \( i \)-th and \( j \)-th nodes are connected by an edge, and \(\mathbf{A}_{i,j} = 0\) otherwise.

Given a source graph \( \mathcal{G}_s = (\mathcal{V}_s, \mathcal{E}_s) \) with \( m \) labeled nodes and a target domain graph \( \mathcal{G}_t = (\mathcal{V}_t, \mathcal{E}_t) \) with \( n \) unlabeled nodes, we assume that both graphs share the same feature and label space but have different marginal distributions,  i.e., \( P(\mathcal{X}_s) \neq P(\mathcal{X}_t) \) but \( P(\mathcal{Y}_s \mid \mathcal{X}_s) = P(\mathcal{Y}_t \mid \mathcal{X}_t) \). 
In this paper, our goal is to develop a GNN-based classifier that can predict the node labels \( \{ y_i^t \}_{i=1}^{n} \) in the target domain, where \( y_i^t \in \mathcal{Y}_t \). We structure the model as a composition \( g \circ \phi \), where the mapping \( \phi: \mathcal{X} \rightarrow \mathcal{H} \) transforms the features into a latent space \(\mathcal{H}\), and the function \( g: \mathcal{H} \rightarrow \mathcal{Y} \) performs classification in the label space \(\mathcal{Y}\). 

 
 \section{Methodology}

In this section, we begin by presenting target domain structural smoothing tailored for UGDA. Following this, we provide a theoretical analysis showing how target error is influenced by model smoothness, demonstrating the method's effectiveness in enhancing knowledge transfer. Lastly, we offer a detailed analysis of the model's complexity.

 \subsection{Target Domain Structural Smoothing}
 As previously discussed, even minor structural differences between the source and target domains can significantly alter node representations. To counteract this, we introduce a straightforward yet powerful approach called target domain structural smoothing, designed to mitigate the effects of structural shifts. The key concept is to enhance the smoothness of node representations in the target graph by constraining the variations between neighboring nodes, thereby minimizing the adverse impact of motif-related structural changes on the model's predictive performance.
 
 The overall process comprises two main steps: neighboring node generation and Laplacian smoothness constraint. In the first step, semantically similar nodes are sampled as neighbors for each node. 
The sampling method is flexible, accommodating various strategies such as random sampling, $k$-hop sampling, or random walk sampling. In this context, we present two specific techniques for neighbor sampling.

 \textbf{k-hop Sampling:} 
In the k-hop sampling, the neighbors of each node are selected by traversing up to $k$ edges in the graph, where $k$ determines the size of the neighborhood. Formally, let \(\mathcal{G} = (\mathcal{V}, \mathcal{E})\) represent an undirected graph, where \(\mathcal{V}\) is the set of nodes and \(\mathcal{E}\) is the set of edges. For a given node \(v_i\), its $k$-hop neighbors \(\mathcal{N}_k(v_i)\) are defined as:
\begin{equation}
	\mathcal{N}_k(v_i) = \{ v_j \in \mathcal{V} \mid \text{shortest\_path}(v_i, v_j) \leq k \},
\end{equation}
where \(\text{shortest\_path}(v_i, v_j)\) denotes the shortest distance between nodes \(v_i\) and \(v_j\).

\textbf{Random Walk Sampling:} The effectiveness of k-hop sampling is highly dependent on the parameter $k$. An inappropriate $k$ can compromise the distinguishability of the learned embeddings. 
For example, a large $k$ might lead to different nodes sharing identical subgraphs, causing their embeddings to collapse into a single point, which is a characteristic of over-smoothing in GNNs.

Instead, the random walk sampling enables flexible exploration of the graph structure during sampling, helping to avoid over-smoothing and maintain meaningful distinctions between nodes~\cite{nikolentzos2020random}. Formally, 
for a given node \(v_{i} \in \mathcal{V}\), the set of neighboring nodes \(\mathcal{N}_r(v_{i})\) is defined as:
\begin{equation}
	\mathcal{N}_r(v_i) = \left\{ v_j \in \mathcal{V} \mid \text{RW}_{\lambda}(v_i, v_j) = \text{True} \right\},
\end{equation}
where $\text{RW}_{\lambda}(v_i, v_j)$ indicates if node $v_j$
is visited during a random walk of length $\lambda$ starting at node $v_i$. 

Next, let \( \mathbf{A} \) be the original adjacency matrix of graph $\mathcal{G}$. Here, we use random walk sampling as an example to illustrate the subsequent process.
The updated adjacency matrix \( \widetilde{\mathbf{A}} \), reflecting the sampled neighborhood, can be defined as:
\begin{equation}
	\widetilde{\mathbf{A}}_{ij} = 
	\begin{cases} 
		1, & \text{if } v_j \in \mathcal{N}_r(v_i)  \text{ and } (v_i, v_j) \in \mathcal{E} \\ 
		0, & \text{otherwise}.
	\end{cases}
\end{equation}

This updated matrix \(\widetilde{\mathbf{A}}\) ensures that only edges between node \(v_i\) and its sampled neighbors \(\mathcal{N}_r(v_i)\) are retained, thereby emphasizing relevant connections within the graph.

Finally, we apply a smoothness constraint for each node to ensure consistency among neighboring nodes in target graph. In this context, we focus on applying this constraint solely to the target graph, as explained in \textbf{Remark 2}.  Let \( f(\mathbf{x}_i) \) denote the feature representation of node \( i \), and \( d_i \) represent the degree of node \( i \). The Laplacian smoothing regularization loss \(\mathcal{L}_{\text{SR}}\) is defined as follows:
\begin{equation}
	\mathcal{L}_{\text{SR}} = \frac{1}{2} \sum_{i} \sum_{j} 	\widetilde{\mathbf{A}}_{ij} \left\| \frac{f(\mathbf{x}_i)}{\sqrt{d_i}} - \frac{f(\mathbf{x}_j)}{\sqrt{d_j}} \right\|^2.
\end{equation}

This constraint minimizes feature differences between nodes and their neighbors, normalized by the square root of their degrees. This normalization ensures that high-degree nodes do not overly dominate the smoothing process.

 \subsection{Target Risk Bound with Model Smoothness}
 In this part, we show that how the target risk is bounded above by the model's smoothness, providing a theoretical foundation for the effectiveness of our approach.
Formally, we define the model smoothness~\cite{rosca2020case} and total variation distance (TVD)~\cite{villani2009optimal}.

\textbf{Definition 1.} \textit{Model Smoothness: } A model $f$ is $(k,r)$-cover with $\Phi$ smoothness on graph $\mathcal{G}$, if
\begin{equation}
\mathbb{E}_\mathcal{G} \left[ \sup_{d(i,j) \leq k, \|\mathbf{x}_i - \mathbf{x}_j\|_\infty \leq r} \left| f(\bm{\theta}, \mathbf{x}_i) - f(\bm{\theta}, \mathbf{x}_j) \right| \right] \leq \Phi,
\end{equation}
where `sup' refers to the supremum, $k$ is the maximum distance between neighboring nodes, and $r$ denotes the maximum difference in the feature space between node features, and $\bm{\theta}$ is the model parameters.
It should note that a lower  \(\Phi\) value indicates a higher level of model smoothness.

\textbf{Definition 2.} \textit{Total Variation Distance: }
Given two graphs $\mathcal{G}_{1}$ and $\mathcal{G}_{2}$, $\mathcal{G}_{1}(v)$ and $\mathcal{G}_{2}(v)$ represent the attribute distributions of node \( v \) on graphs $\mathcal{G}_{1}$ and $\mathcal{G}_{2}$ respectively. The total variation distance is defined as:
\begin{equation}
	\text{TVD}(\mathcal{G}{1}, \mathcal{G}{2}) = \frac{1}{2} \sum_{v \in V} \left| \mathcal{G}{1}(v) - \mathcal{G}{2}(v) \right|,
\end{equation}
where it measures the maximum difference between the attribute distributions of two nodes on the graph.

As we proceed, let's assume that the graphs \(\mathcal{G}_1\) and \(\mathcal{G}_2\) from the source and target domains have a compact support, denoted as \(\mathcal{X} \subseteq \mathbb{R}^d\), where $d$ is the feature dimension. This implies there exists a constant \(\Gamma > 0\) such that for any two nodes \(\mathbf{x}_i, \mathbf{x}_j \in \mathcal{X}\),  the difference between their features, \(\|\mathbf{x}_i - \mathbf{x}_j\|\), is also constrained by \(\Gamma\).
Consider \(\mathcal{L}(f(\mathbf{x}), y)\) as the loss function, which is continuous and differentiable. Here, \(f(\mathbf{x})\) is the output of the GNNs predicting the node label, and \(y\) is the ground truth of node. We define the expected risk for the model \(f\) across the distribution \(\mathbb{S}\) as \(\mathcal{E}_{\mathbb{S}}(f) = \mathbb{E}_{\{\mathbf{x}, y\} \sim \mathbb{S}} \left[\mathcal{L}(f(\mathbf{x}), y)\right]\). Additionally, we assume that the loss is bounded, i.e., \(0 \leq \mathcal{L}(f(\mathbf{x}), y) \leq \Upsilon\). By combining our understanding of model smoothness and the concept of TVD, we can establish the following theorem.

\begin{theorem}
	Consider two distributions $\mathbb{S}$ and $\mathbb{T}$, if a model $f$ is $(k,2r)$-cover with  $\Phi$ smoothness over distributions $\mathbb{S}$ and $\mathbb{T}$, then with probability at least $1- \xi$, the following holds:
	\begin{equation}
		\begin{aligned}
			\mathcal{E}_{\mathbb{T}}(f) &\leq \mathcal{E}_{\mathbb{S}}(f)+ 2 \mathrm{TVD}(\mathbb{S}, \mathbb{T}) +  \Phi_{\mathbb{S}}+\Phi_{\mathbb{T}}+K,
		\end{aligned} 
	\label{th1}
	\end{equation}
	where
\begin{equation}
	K = \frac{\Upsilon Z}{\sqrt{m}} + \frac{\Upsilon Z}{\sqrt{n}} + \Upsilon \sqrt{\frac{\log (1/ \xi)}{2m}},
	\label{k}
\end{equation}
	and
	\begin{equation}
	Z = \sqrt{(2d)^{\frac{2 \Phi^2 \Gamma}{r^2} + 1} \log 2 + 2 \log (1/ \xi)}.
\end{equation}
\begin{proof}
	Refer to the Appendix for the details.
\end{proof}
\end{theorem}

In this theorem, $K$ and $Z$ are intermediate variables, $\Gamma$ and $\Upsilon$ are constants specific to the given dataset, and $m$ and $n$ denote the number of nodes in the source and target domains, respectively.
It is important to note that, unlike previous methods that derive upper bounds for target error~\cite{you2023graph,liu2024rethinking}, we propose that the target risk \(\mathcal{E}_{\mathbb{T}}(f)\) is constrained by the source risk \(\mathcal{E}_{\mathbb{S}}(f)\), domain discrepancy \(\text{TVD}(\mathbb{S}, \mathbb{T})\), and the model's smoothness \(\Phi\). This relationship is encapsulated in Theorem 1, which provides a more comprehensive understanding of the factors impacting the target risk. Therefore, by reducing model's smoothness \(\Phi\), we can effectively decrease the target error, offering a novel perspective for addressing the UGDA problem. It is also the core of our proposed TDSS approach.

\vspace{4pt}
\noindent
\textbf{Remark 1: (Why not directly increase GNNs depth?)} 
Some readers may wonder why we do not simply increase the depth of GNNs to enhance model smoothness. While adding more layers can increase smoothness, it often results in over-smoothing, which leads to the loss of critical node-specific features. In contrast, our proposed TDSS provides a better balance by enhancing smoothness while preserving local structural details, making it ideal for improving model robustness in UGDA tasks.
Additionally, TDSS aligns more closely with the mathematical formulation of the target risk bound (i.e., Eq. (\ref{th1})), as it directly influences $\Phi$, the smoothness parameter we aim to optimize. This alignment ensures that our theoretical insights translate into tangible improvements in model' generalization capabilities.

\vspace{4pt}
\noindent
\textbf{Remark 2: (Target or source domain?) } 
Since UGDA involves two distributions, \(\mathbb{S}\) (source domain) and \(\mathbb{T}\) (target domain), one might intuitively consider smoothing both the source and target graphs simultaneously. However, this approach can be computationally expensive, necessitating further optimization. In fact, the target domain \(\mathbb{T}\) generally requires greater smoothness than the source domain \(\mathbb{S}\) because the source domain benefits from labeled data, enabling more stable feature representations. Conversely, the target domain lacks supervision, making it more susceptible to noise and variability, thus requiring enhanced smoothness for effective generalization. Thus, we have \(\Phi_{\mathbb{S}} < \Phi_{\mathbb{T}}\). As a consequence, Eq. (\ref{th1}) can be reformulated as:
\begin{equation}
	\begin{aligned}
		\mathcal{E}_{\mathbb{T}}(f) &\leq \mathcal{E}_{\mathbb{S}}(f)  + 2 \mathrm{TVD}(\mathbb{S}, \mathbb{T}) + 2\Phi_{\mathbb{T}}+K,
	\end{aligned}
\label{re2}
\end{equation}
where $K$ is defined in Eq.~(\ref{k}). This approach reduces computational costs while minimizing the target error.


\vspace{4pt}
\noindent
 \textbf{Remark 3: (Why total variation distance?)}
 In proving target error, we opt to use TVD instead of MMD or other measures due to its robust and interpretable assessment of distributional discrepancy. TVD evaluates the maximum difference between probabilities assigned by two distributions, making it particularly suitable for scenarios where absolute differences are critical, and is favored for its convenience in theoretical derivations~\cite{devroye2018total}.
On the contrary, MMD captures higher-order moment differences but relies on kernel choices and large sample sizes for convergence, adding complexity and computational overhead.
 Thus, we use TVD to analyze the new target error bound $\mathcal{E}_{\mathbb{T}}(f)$.
While in fact, with suitable kernel functions, MMD can approximate an upper bound for TVD:
 \begin{equation}
 	\mathrm{TVD}(\mathbb{S}, \mathbb{T}) \leq \text{MMD}_k(\mathbb{S}, \mathbb{T}) \cdot C_k,
 	\label{re3}
 \end{equation}
 where \(C_k\) is a constant related to the kernel function. Consequently, these measures of domain discrepancy can be regarded as equivalent under mild conditions, thereby preserving the correctness of our derivations regardless of the metric chosen. 
Furthermore, by combining Eq. (\ref{re2}) and Eq. (\ref{re3}), the new target error can be rewritten as:
\begin{equation}
	\begin{aligned}
		\mathcal{E}_{\mathbb{T}}(f) &\leq \underbrace{\mathcal{E}_{\mathbb{S}}(f)}_{\mathcal{L}_{\text{GC}}} + \underbrace{2 \mathrm{MMD}_k(\mathbb{S}, \mathbb{T})\cdot C_k}_{\mathcal{L}_{\text{DA}}}  + \underbrace{2\Phi_{\mathbb{T}}}_{\mathcal{L}_{\text{SR}}}+ K,
	\end{aligned}
\end{equation}
 where $K$ is defined in Eq.~(\ref{k}). Interestingly, we observe that the target error is primarily constrained by three factors (i.e., source error, the MMD distance between domains, and model smoothness of target domain), each of which aligns closely with our optimization objectives in Eq. (\ref{loss}). This alignment indicates a strong consistency between our theoretical framework and the practical optimization goals, further reinforcing the validity of our approach.

\subsection{Overall Optimization}
The overall training process of our proposed TDSS consists of three main components:
\begin{equation}
	\mathcal{L} =  \mathcal{L}_{\text{GC}} + \alpha\mathcal{L}_{\text{DA}} + \beta  \mathcal{L}_{\text{SR}},
	\label{loss}
\end{equation}
where the first term, \(\mathcal{L}_{\text{GC}}\), represents the GNNs classifier loss. In this paper, we employ the state-of-the-art A2GNN~\cite{liu2024rethinking} to optimize this component. The \(\mathcal{L}_{\text{DA}}\) is the domain alignment loss, where we utilize the MMD method to align the source and target domains, aiming to minimize domain discrepancies. Finally, \(\mathcal{L}_{\text{SR}}\) is the smoothness loss, designed to ensure node smoothness by constraining feature variations between neighboring nodes. The \(\alpha\) and \(\beta\) act as trade-offs to balance the contributions of the domain alignment and smoothness losses, respectively.


\vspace{4pt}
\noindent
\textbf{Complexity Analysis.}
In this section, we analyze the time complexity of TDSS, using the random walk sampling method as an example.
Firstly, for each node performing \( \gamma \) random walks with a walk length of \( \lambda \), the time complexity is \( O(n\gamma \lambda) \), where $n$ is the node number in target graph. The values of \(\gamma\) and \(\lambda\) are chosen to be small, thus not incurring significant computational overhead.  Secondly, for each node calculating the smoothing constraint with its random walk sampled neighbors, assuming that each node samples on average \(\rho\) neighbors (i.e., \(\rho \leq \gamma \lambda \)), the time complexity is \( O(n\rho d) \), where \(d\) is the number of features. Therefore, the overall time complexity is \( O(n \gamma \lambda + n \rho d) \).

\section{Experiment}
In this section, we conduct experiments on three real-world datasets to address the following five research questions:
 \textbf{(RQ1)}: How does the proposed method TDSS perform when compared to state-of-the-art baseline methods?
 \textbf{(RQ2)}: How is the effectiveness of the proposed model evaluated across different backbone GNNs architectures?
\textbf{(RQ3)}: Which part of the model primarily contributes to the effective prediction of optimal UGDA?
\textbf{(RQ4)}: How do the hyper-parameters affect the performance of the proposed approach?
 \textbf{(RQ5)}: How about the intuitive effect of TDSS?

\subsection{Experimental Setups}
\subsubsection{Datasets}
We conduct experiments utilizing three real-world graphs from the ArnetMiner dataset~\cite{tang2008arnetminer}: \textbf{ACMv9 (A)}, \textbf{Citationv1 (C)}, and \textbf{DBLPv7 (D)}. To address the discrepancies in node attributes across these graphs, we integrate their attribute sets and standardized the attribute dimension to 6775, as elaborated in \cite{qiao2023semi}. Each node signifies a paper, while each edge denotes a citation between two papers. Our objective is to categorize all the papers into five distinct research topics: Databases, Artificial Intelligence, Computer Vision, Information Security, and Networking. This study focuses on six key transfer tasks: \textbf{A $\rightarrow$ C}, \textbf{A $\rightarrow$ D}, \textbf{C $\rightarrow$ A}, \textbf{C $\rightarrow$ D}, \textbf{D $\rightarrow$ A}, and \textbf{D $\rightarrow$ C}. The statistics of three datasets are summarized in Table~\ref{data}.

\begin{table}[t]
	\renewcommand{\arraystretch}{1.2}
	    \caption{Statistics of the three real-world graphs. Note: ‘\#’ denotes ‘number of’, ‘Attr.’ refers to attributes.}
	\resizebox{0.48\textwidth}{!}{
		\centering
		\begin{tabular}{cccccc}
	\bottomrule
			Graph & \#Nodes & \#Edges & \#Attr. & \#Labels & \#Density\\
			\hline \hline
			ACMv9 (\textbf{A}) & 9,360 & 15,556 & 6,775 & 5 & 0.00036 \\
			Citationv1(\textbf{C})  & 8,935 & 15,098 & 6,775 & 5 & 0.00038 \\
			DBLPv7(\textbf{D})  & 5,484 & 8,117 & 6,775 & 5 & 0.00054 \\
			\bottomrule
	\end{tabular}}
    \label{data}
\end{table}

\begin{table*}[t]
	\centering
		\caption{Node classification performance comparisons on six cross-domain tasks.  The highest scores are highlighted in \textbf{bold}, and the second-highest scores are \underline{underlined}. $*$ indicates that TDSS significantly outperforms A2GNN at the 0.05 level.}
	\renewcommand\arraystretch{1.3}
	\resizebox{0.99\linewidth}{!}{%
		\begin{tabular}{l|cccccccccccc|cc}
			\toprule
			\multirow{2}{*}{Methods} & \multicolumn{2}{c}{\textbf{A $\to$ C}} & \multicolumn{2}{c}{\textbf{A $\to$ D}} & \multicolumn{2}{c}{\textbf{C $\to$ A}} & \multicolumn{2}{c}{\textbf{C $\to$ D}} & \multicolumn{2}{c}{\textbf{D $\to$ A}} & \multicolumn{2}{c|}{\textbf{D $\to$ C}} & \multicolumn{2}{c}{Avg.} \\
			& Micro-F1 & Macro-F1 & Micro-F1 & Macro-F1 & Micro-F1 & Macro-F1 & Micro-F1 & Macro-F1 & Micro-F1 & Macro-F1& Micro-F1 & Macro-F1 & Micro-F1 & Macro-F1 \\
			\hline 			\hline
			DeepWalk& 0.2105 & 0.1772 & 0.2594 & 0.1987 & 0.2194 & 0.1933 & 0.2257 & 0.1751 & 0.2623 & 0.1983 & 0.2946 & 0.2276 & 0.2453 & 0.1956 \\
			node2vec & 0.2989 & 0.2584 & 0.2454 & 0.1950 & 0.2176 & 0.1799 & 0.2895 & 0.2498 & 0.2861 & 0.2205 & 0.2116 & 0.1622 & 0.2582 & 0.2108 \\
			ANRL & 0.3031 & 0.2093 & 0.2954 & 0.2333 & 0.3184 & 0.2204 & 0.2590 & 0.2271 & 0.2956 & 0.1912 & 0.2599 & 0.1825 & 0.2886 & 0.2099 \\ \hdashline
			GAT & 0.5713 & 0.4364 & 0.5380 & 0.4136 & 0.5037 & 0.4214 & 0.5585 & 0.4525 & 0.5293 & 0.4395 & 0.5552 & 0.5004 & 0.5427 & 0.4436 \\ 
			GSAGE & 0.7140 & 0.6914 & 0.6482 & 0.6180 & 0.6522 & 0.6469 & 0.6996 & 0.6686 & 0.5922 & 0.5731 & 0.6785 & 0.6490 & 0.6641 & 0.6379 \\
			SGC & 0.7740 & 0.7213 & 0.6913 & 0.6254 & 0.7231 & 0.6292 & 0.7380 & 0.6693 & 0.6332 & 0.5394 & 0.7231 & 0.6292 & 0.7138 & 0.6290 \\
			GCN & 0.7738 & 0.7478 & 0.6905 & 0.6529 & 0.7058 & 0.7039 & 0.7417 & 0.7137 & 0.6335 & 0.5942 & 0.7417 & 0.6979 & 0.7145 & 0.6824 \\ \hdashline
			UDAGCN & 0.7215 & 0.6033 & 0.6695 & 0.6483 & 0.5816 & 0.5589 & 0.7177 & 0.6946 & 0.5816 & 0.5589 & 0.7328 & 0.6112 & 0.6675 & 0.5986 \\
			GRADE & 0.7604 & 0.7252 & 0.6822 & 0.6303 & 0.6372 & 0.5935 & 0.7395 & 0.7002 & 0.6372 & 0.5935 & 0.7432 & 0.6932 & 0.6999& 0.6551 \\
			CDNE & 0.7876 & 0.7683 & 0.7158 & 0.6924 & 0.6962 & 0.7045 & 0.7436 & 0.7134 & 0.6962 & 0.7045 & 0.7888 & 0.7736 & 0.7380 & 0.7362 \\
			AdaGCN & 0.7932 & 0.7651 & 0.7504 & 0.7139 & 0.6967 & 0.6947 & 0.7559 & 0.7234 & 0.6967 & 0.6947 & 0.7820 & 0.7422 & 0.7458 & 0.7254 \\
			StruRW& 0.7735 & 0.7207 & 0.6910 & 0.6251 & 0.6781 & 0.5977 & 0.7381 & 0.6689 & 0.6327 & 0.5382 & 0.7241 & 0.6294 & 0.7016 & 0.6234 \\
			PairAlign& 0.7088 & 0.6788 & 0.6591 & 0.6235 & 0.6585 & 0.6509 & 0.7104 & 0.6756 & 0.5934 & 0.5877 & 0.6707 & 0.6461 & 0.6602 & 0.6338 \\
			SpecReg & 0.8055 & 0.7883 & 0.7593 & 0.7398 & 0.7101 & 0.7234 & 0.7574 & 0.7364 & 0.7101 & 0.7234 & 0.7904 & 0.7778 & 0.7556 & 0.7501 \\
			ACDNE & 0.8175 & 0.8009 & 0.7624 & 0.7359 & 0.7129 & 0.7264 & 0.7721 & 0.7574 & 0.7129 & 0.7264 & 0.8014 & 0.7883 & 0.7513& 0.7425 \\
			A2GNN & 0.8275 & 0.8086 & 0.7784 & 0.7518 & \underline{0.7463} & \underline{0.7631}& \underline{0.7828} & \underline{0.7590} & 0.7344 & 0.7494 & 0.8109 & 0.7884 & 0.7795 & 0.7672 \\ \hdashline  \rowcolor{tan}
						\textbf{TDSS}$_\text{k-hop}$ &\underline{0.8294}  &\underline{0.8131}  &\underline{0.7846}  &\underline{0.7611}  &0.7433 &0.7573 &0.7713 & 0.7527 &\underline{0.7399} &\underline{0.7573}  &\underline{0.8177} &\underline{0.8011}  &\underline{0.7815}  & \underline{0.7765} \\  \rowcolor{tan}
			\textbf{TDSS}$_\text{RW}$ & \textbf{0.8320$^{*}$} & \textbf{0.8140$^{*}$} & \textbf{0.7954$^{*}$} & \textbf{0.7813$^{*}$} & \textbf{0.7694$^{*}$}& \textbf{0.7833$^{*}$} & \textbf{0.7896$^{*}$} & \textbf{0.7755$^{*}$} & \textbf{0.7489$^{*}$} & \textbf{0.7646$^{*}$} & \textbf{0.8229$^{*}$} & \textbf{0.8075$^{*}$} & \textbf{0.7954} & \textbf{0.7906} \\
			\bottomrule
	\end{tabular}}
	\label{main}
\end{table*}
\subsubsection{Baselines}
For performance evaluation, we compare TDSS with three categories of baseline methods:
(1) Unsupervised graph learning:  DeepWalk~\cite{perozzi2014deepwalk}, node2vec~\cite{grover2016node2vec}, and ANRL~\cite{zhang2018anrl} learn node embeddings without supervision and then evaluate target graph representations using a classifier trained on the source graph.
(2) Source-only GNNs:  GAT~\cite{velickovic2017graph}, GSAGE~\cite{hamilton2017inductive}, SGC~\cite{wu2019simplifying}, and GCN~\cite{kipf2016semi}, are trained on the source graph in an end-to-end fashion, allowing direct application to the target graph.
(3) Graph domain adaptation methods: CDNE~\cite{shen2020network}, AdaGCN~\cite{dai2022graph}, ACDNE~\cite{shen2020adversarial}, UDAGCN~\cite{wu2020unsupervised},  GRADE~\cite{wu2023non}, StruRW~\cite{liu2023structural}, PairAlign~\cite{liu2024pairwise} SpecReg~\cite{you2023graph} and A2GNN~\cite{liu2024rethinking} are designed specifically to handle graph domain adaptation.

\subsubsection{Implementation Details} 
Consistent with previous studies~\cite{shen2020network,liu2024rethinking}, we use Micro-F1 and Macro-F1 scores to evaluate model performance. Node features are set to a dimensionality of 128, and the learning rate was fine-tuned within the set \{0.01, 0.03, 0.05\} for optimal performance. Domain alignment is achieved using the MMD method with a Gaussian kernel~\cite{filippone2008survey}. To prevent overfitting, a dropout rate of 0.5 is applied. We tune the \(\alpha\) and \(\beta\) parameters within the range \([0, 1]\). All experiments are conducted using PyTorch, ensuring consistency with baseline models by adhering to the implementation details from their original papers.

\subsection{Experiment Results}
\subsubsection{Performance Comparison (RQ1)}

We present the performance of TDSS compared with baselines for node classification, as shown in Table~\ref{main}. From the results, we can observe:

(1) Graph domain adaptation methods outperform vanilla graph neural networks and hypothesis transfer approaches, underscoring the importance of addressing domain discrepancies. Vanilla GNNs struggle with domain shifts due to the assumption of identical training and test distributions. Hypothesis transfer methods adapt to new domains but fail to explicitly address graph structure discrepancies. In contrast, graph domain adaptation methods model domain shifts and adapt graph representations, leading to superior performance and highlighting the need for domain-aware adaptations in graph-based learning tasks.

 (2) The proposed TDSS$_\text{RW}$ model achieves the best performance across six cross-domain node classification datasets, outperforming all baseline methods. On average, TDSS$_\text{RW}$ shows a 2.04\% improvement in Macro-F1 and a 3.05\% improvement in Micro-F1 over the state-of-the-art A2GNN model. These gains are due to the incorporation of model smoothness, which mitigates the impact of local structural changes. The random walk-based Laplacian smoothing constraint in TDSS$_\text{RW}$ enables better capture of the intrinsic graph structure and enhances robustness against noise, resulting in more accurate node classification. Moreover, the TDSS\(_{\text{k-hop}}\) model underperforms compared to TDSS\(_{\text{RW}}\), likely due to over-smoothing during the optimization process. In k-hop sampling, nodes may share similar subgraphs as the hop count increases, leading to indistinguishable embeddings and a loss of critical node-specific features, which reduces the model's effectiveness.

 (3) We observe that the improvement in the \textbf{C $\to$ A} setting is the most significant. This enhancement may be attributed to the similarity in the number of nodes and edges between the Citationv1 and ACMv9 datasets. Such similarity enables the random walk-based smoothing constraint method to more effectively leverage the structural and attribute correspondences, resulting in smoother adaptation and higher classification accuracy in this specific transfer setting. 
 
\begin{table}[t]
	\centering
		\caption{Node classification performance on various GNNs for unsupervised domain adaptation.}
			    \vspace{-5pt}
	\renewcommand\arraystretch{1.3}
	\resizebox{\linewidth}{!}{%
		\begin{tabular}{cc|cccccc}
			\bottomrule
			\multicolumn{2}{c|}{\multirow{2}{*}{Backbone}} & \multicolumn{2}{c|}{GCN}                     & \multicolumn{2}{c|}{GAT}                     & \multicolumn{2}{c}{SGC} \\
			\multicolumn{2}{c|}{}                      & Original & \multicolumn{1}{c|}{ Ours} & Original  & \multicolumn{1}{c|}{ Ours } & Original  &  Ours  \\ \hline \hline
			\multirow{2}{*}{\textbf{A $\to$ C}}        & Mi-F1       & 0.7813          & \cellcolor{tan}\textbf{0.7930}                               & 0.7545         & \cellcolor{tan}\textbf{0.7996}                            & 0.7632           & \cellcolor{tan}\textbf{0.7765}          \\  
			& Ma-F1       & 0.7598          & \cellcolor{tan}\textbf{0.7758}                              & 0.7360         & \cellcolor{tan}\textbf{0.7748}                                & 0.6983        &\cellcolor{tan} \textbf{0.7224}         \\  
			\multirow{2}{*}{\textbf{A $\to$ D}}        & Mi-F1       & 07055          & \cellcolor{tan}\textbf{0.7255}                                & 0.7213          & \cellcolor{tan}\textbf{0.7490  }                           & 0.7535         & \cellcolor{tan}\textbf{0.7644}            \\
			& Ma-F1       & 0.6750         & \cellcolor{tan}\textbf{0.6967}                               & 0.6861         & \cellcolor{tan}\textbf{0.7218}                              & 0.6852          & \cellcolor{tan}\textbf{0.7083}         \\ 
			\multirow{2}{*}{\textbf{C $\to$ A}}        & Mi-F1       & 0.7150          & \cellcolor{tan}\textbf{0.7230}                                 & 0.7233          & \cellcolor{tan}\textbf{0.7425}                                 & 0.6657         & \cellcolor{tan}\textbf{0.6818}            \\
			& Ma-F1       & 0.7184          & \cellcolor{tan}\textbf{0.7298}                                 & 0.7291          & \cellcolor{tan}\textbf{0.7529}                                 & 0.5954         & \cellcolor{tan}\textbf{0.6394}         \\ 
			\multirow{2}{*}{\textbf{C $\to$ D}}        & Mi-F1       & 0.7583          &\cellcolor{tan} \textbf{0.7621}                             & 0.7403          & \cellcolor{tan}\textbf{0.7740}                              & 0.7485        & \cellcolor{tan}\textbf{0.7546}        \\
			& Ma-F1       & 0.7343        & \cellcolor{tan}\textbf{0.7405}                              & 0.7153         &\cellcolor{tan} \textbf{0.7529}                               & 0.6791          & \cellcolor{tan}\textbf{0.7029}        \\ 
			\multirow{2}{*}{\textbf{D $\to$ A}}        & Mi-F1       & 0.6568          & \cellcolor{tan}\textbf{0.6709}                               & 0.6795        & \cellcolor{tan}\textbf{0.7157}                               & 0.6252          & \cellcolor{tan}\textbf{0.6467}          \\
			& Ma-F1       & 0.6465         & \cellcolor{tan}\textbf{0.6700}                              & 0.6862        & \cellcolor{tan}\textbf{0.7136}                               & 0.5364          & \cellcolor{tan}\textbf{0.5549}          \\ 
			\multirow{2}{*}{\textbf{D $\to$ C}}        & Mi-F1       & 0.7487          & \cellcolor{tan}\textbf{0.7632}                               & 0.7652          & \cellcolor{tan}\textbf{0.7902}                               & 0.7374         & \cellcolor{tan}\textbf{0.7637}         \\
			& Ma-F1       & 0.7187         & \cellcolor{tan}\textbf{0.7339}                            & 0.7347         & \cellcolor{tan}\textbf{0.7756}                             & 0.6380          & \cellcolor{tan}\textbf{0.6732}         \\ 	\bottomrule
	\end{tabular}}
	\label{backone}
\end{table}
\subsubsection{Results for Various GNNs (RQ2).} 
Since TDSS\footnote{To simplify notation, we use TDSS to refer to TDSS\(_{\text{RW}}\) in the following experiments.
} is designed to be model-agnostic and seamlessly integrable into existing GNNs frameworks, it can be effectively integrated not only into state-of-the-art models like A2GNN but also into standard GNNs. To validate its broad applicability, we conduct experiments with GCN, GAT, and SGC. As shown in Table~\ref{backone}, the results demonstrate significant performance gains with our method compared to the original models without the smoothing constraint.
Overall, these findings clearly indicate that our proposed TDSS approach enhances the generalization capabilities of various GNNs models. The consistent improvements across different tasks and backbones underscore the effectiveness and versatility of TDSS in advancing graph domain adaptation techniques.

\begin{figure}[t]
	\centering
	\begin{subfigure}[!]{0.23\textwidth}
		\includegraphics[width=\textwidth]{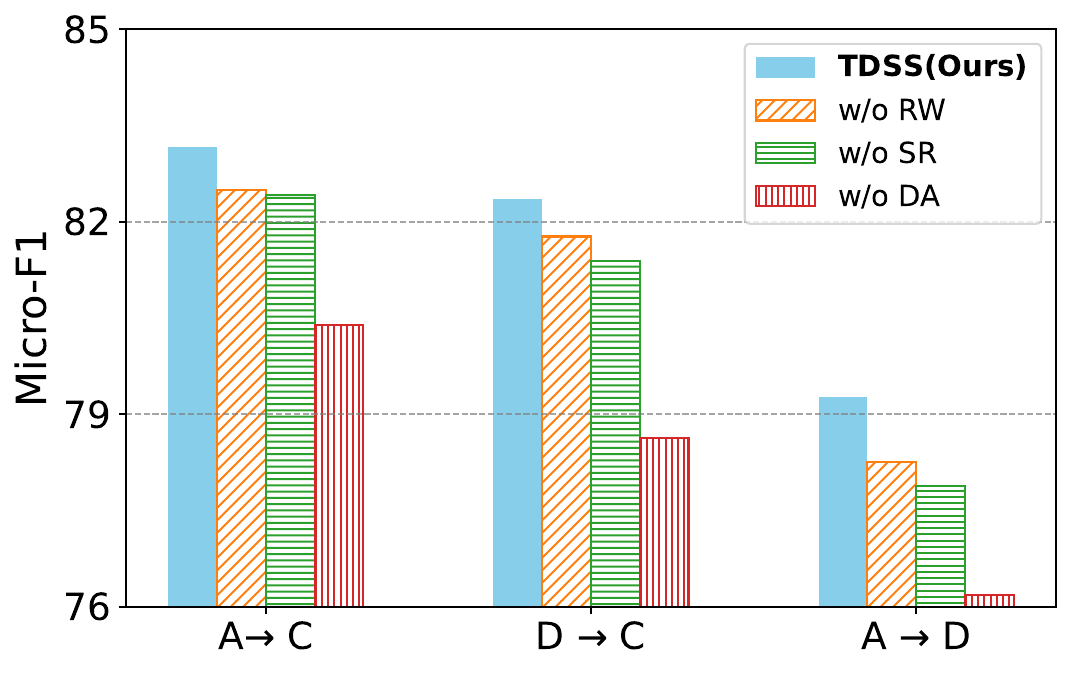}
	\end{subfigure} 
	\begin{subfigure}[!]{0.23\textwidth}
		\includegraphics[width=\textwidth]{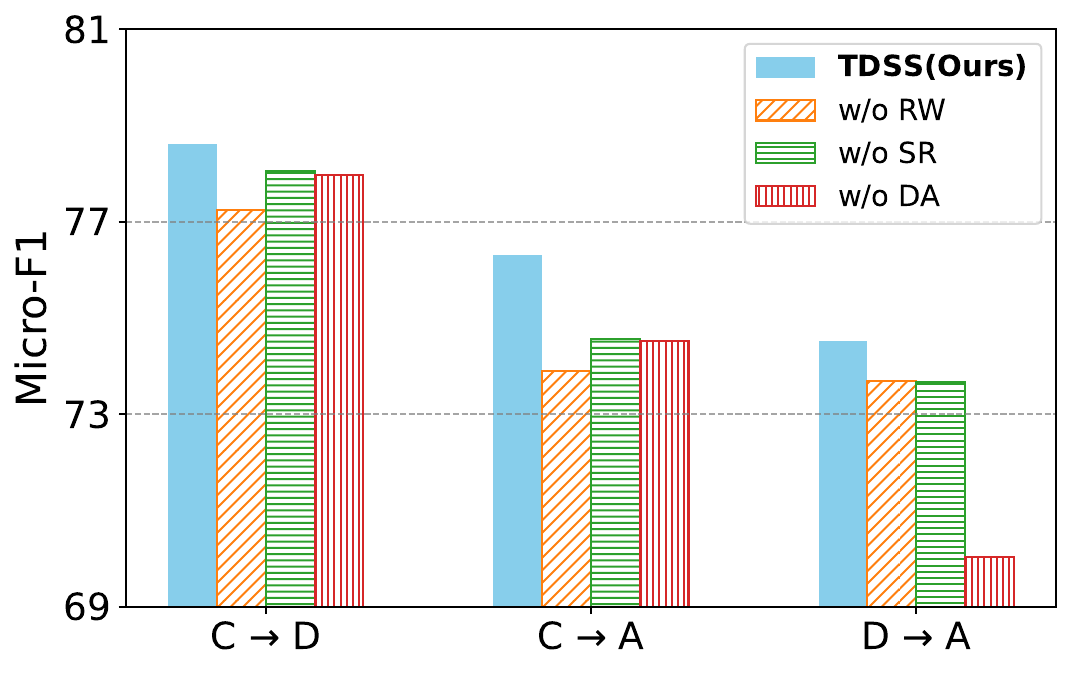}
	\end{subfigure}
	\caption{Classification Micro-F1 comparisons between TDSS variants on six cross-domain tasks.}
	\label{ablation}
\end{figure}

\subsubsection{Ablation Study (RQ3).} 
To investigate the specific contributions of different components, we conduct an in-depth analysis and ablation studies on TDSS. The results are shown in Figure~\ref{ablation}, and the variants of TDSS are: \textbf{(1) w/o DA:} which eliminates the alignment loss to assess the impact of domain alignment (i.e., setting \(\alpha = 0\)). \textbf{(2) w/o SR:} which eliminates the smooth regularization loss to evaluate the impact of model smoothness (i.e., \(\beta = 0\)).
\textbf{(3) w/o RW:} which removes the random walk sampling mechanism and reverts to 1-hop neighborhood sampling.

First, we observe that removing specific modules consistently leads to a decrease in F1 scores, highlighting the essential role that various model components play in determining overall performance. Notably, in the majority of transfer tasks, the removal of MMD alignment (w/o DA) causes the most significant performance drop, emphasizing the critical importance of explicit domain alignment in minimizing distributional differences between two domains. 

Further analysis of the comparison between TDSS and its variants on {\textbf{C $\to$ D}} and {\textbf{C $\to$ A}} reveals a key insight: the ``w/o RW" variant shows inconsistent performance compared to other tasks, likely due to insufficient smoothness when using 1-hop neighbors for Laplacian constraints.

\begin{figure}[t]
	\centering
	\begin{subfigure}[t]{0.23\textwidth}
		\includegraphics[width=\textwidth]{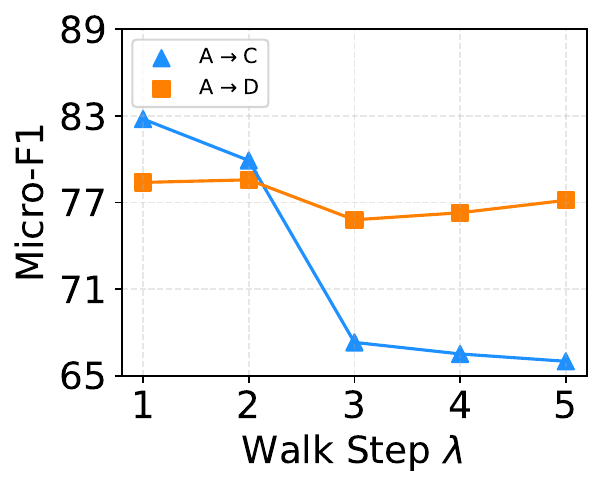}
	\end{subfigure}
	\begin{subfigure}[t]{0.23\textwidth}
		\includegraphics[width=\textwidth]{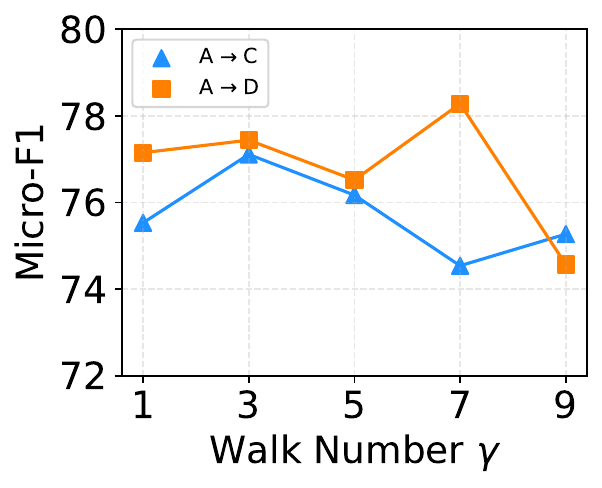}
	\end{subfigure}
	\begin{subfigure}[t]{0.23\textwidth}
		\includegraphics[width=\textwidth]{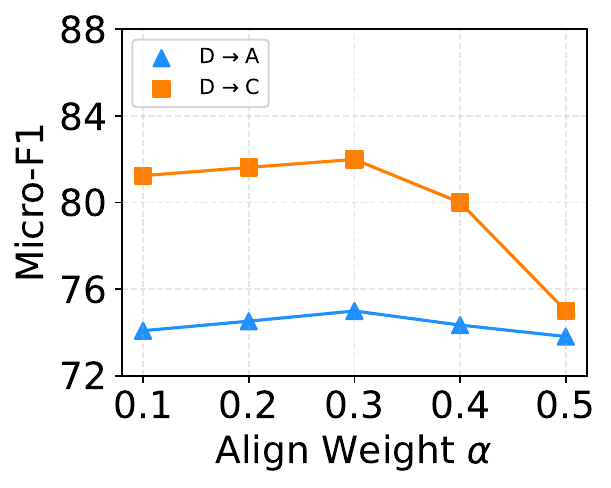}
	\end{subfigure} 
	\begin{subfigure}[t]{0.23\textwidth}
		\includegraphics[width=\textwidth]{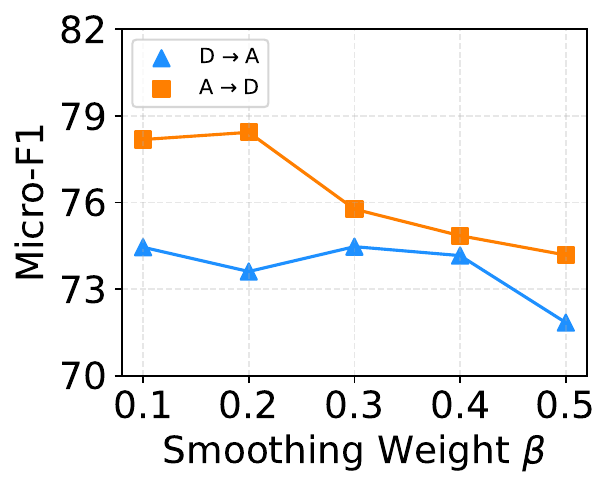}
	\end{subfigure}
	\caption{The Micro-F1 performances of our TDSS w.r.t varying $\lambda$, $\gamma$, $\alpha$, and $\beta$ on different transfer tasks.}
	\label{para}
	\vspace{-5pt}
\end{figure}
\subsubsection{Sensitivity Analysis (RQ4).} 
In this section, we investigate the sensitivity of TDSS to three key components: the random walk mechanism (walk number \(\gamma\) and walk length \(\lambda\)), domain alignment (weight \(\alpha\)), and the smoothing constraint (weight \(\beta\)).
 Due to limited space, we present the experimental results of two transfer tasks per parameter in Figure~\ref{para}. 

$\bullet$ \textbf{Impact of walk length $\lambda$ and walk number $\gamma$.}
Firstly, we investigate the influence of  \(\lambda\) and \(\gamma\). 
When analyzing the walk step \(\lambda\), distinct trends appear in the performance metrics. As \(\lambda\) increases, the Micro-F1 for {\textbf{A $\to$ C}} decreases, suggesting that longer walk steps introduce noise and over-smoothing. For {\textbf{A $\to$ D}}, the Micro-F1 rises initially, peaking at \(\lambda = 2\), before declining, indicating that a moderate walk step captures useful information without noise, emphasizing the need for domain-specific tuning.
For the walk number \(\gamma\), performance slightly drops at higher values, suggesting that a moderate number of walks (e.g., \(\gamma = 3\) to \(\gamma = 5\)) balances information capture and redundancy. Excessive walks result in overlap, reducing performance.
In summary, fine-tuning the walk step and number is essential for optimal UGDA performance, avoiding over-smoothing and redundancy.


$\bullet$ \textbf{Impact of domain aligning weight $\alpha$. }
Secondly, we study the impact  of $\alpha$, a parameter that regulates the intensity of domain aligning. In this analysis, we manipulate the value of $\alpha$, selecting from the set $\{0.1,0.2,0.3,0.4,0.5\}$. 
Notably, we observe that optimal Micro-F1 performance is attained when $\alpha = 0.3$ for both of these transfer tasks. As $\alpha$ continues to increase beyond this point, we observe a decline in Micro-F1 score, suggesting the significance of selecting an appropriate value for $\alpha$ in our model.

$\bullet$ \textbf{Impact of smoothing weight $\beta$.}
Finally, we analyze the impact of the weight \(\beta\), which controls the strength of the smoothing constraint. We vary its values within the set \(\{0.1, 0.2, 0.3, 0.4, 0.5\}\). In the {\textbf{A $\to$ D}} setting, optimal performance is achieved at \(\beta = 0.2\), after which the Micro-F1 score gradually decreases. Similarly, in the {\textbf{D $\to$ A}} transfer scenario, the model reaches its best trade-off at \(\beta = 0.3\). These results suggest that a smaller \(\beta\) is more effective for optimizing the smoothing constraint loss in TDSS.

\begin{figure}[t]
	\vspace{-0pt}
	\centering
	\setlength{\fboxrule}{0.pt}
	\setlength{\fboxsep}{0.pt}
	\fbox{
		\includegraphics[width=\linewidth]{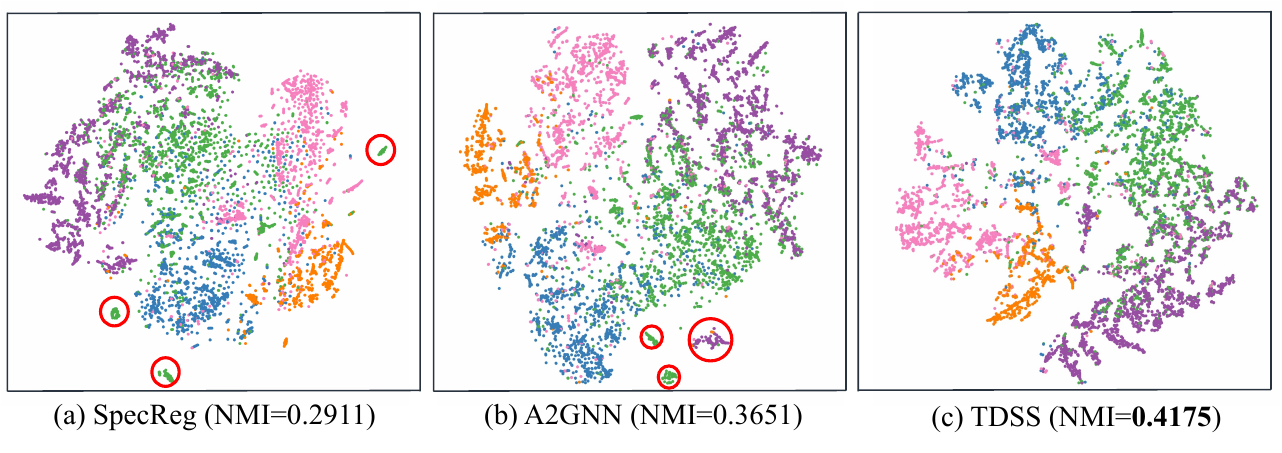}
	}
	\caption{Visualization of learned node embedding on \textbf{D $\to$ A} for different models. We use Normalized
		Mutual Information (NMI)~\cite{zhang2015evaluating} to evaluate the clustering effect, with higher values indicating better performance.
	}
	\label{vis}
\end{figure}

\subsubsection{Visualization Analysis (RQ5).} 
To provide an intuitive understanding of TDSS, we visualize the node representations learned in the target domain for the {\textbf{D $\to$ A}} task, comparing it with three models. Node embeddings are visualized using Stochastic Neighbor Embedding (SNE)~\cite{hinton2002stochastic}, as shown in Fig~\ref{vis}.
TDSS achieves the best clustering performance with an NMI of 0.4175, indicating clear separation between clusters and tighter groupings of nodes. In contrast, SpecReg and A2GNN exhibit more overlap and dispersion, with NMIs of 0.2911 and 0.3651, respectively. The red circles highlight areas where TDSS reduces inter-cluster mixing, demonstrating superior discriminative power and feature representation in complex domains.

\section{Conclusion}

In this paper, we present target-domain structural smoothing, a simple yet effective method for enhancing model transferability in UGDA tasks. By integrating Laplacian smoothing with a neighborhood sampling mechanism, TDSS preserves consistent node representations and strengthens model robustness against structural distribution shifts. Our theoretical analysis underscores the importance of addressing model smoothness for successful domain adaptation. Empirical results on three real-world datasets validate the superiority of TDSS over state-of-the-art baselines, and demonstrate significant improvements across a range of transfer scenarios.

\section{Acknowledgments}
We sincerely thank all the anonymous reviewers for their valuable comments to improve this paper. 
The research work is supported by the National Key Research and Development Program of China under Grant Nos. 2021ZD0113602, the National Natural Science Foundation of China under Grant No. 62176014 and No. 62206266, the Fundamental Research Funds for the Central Universities.

\bibliography{aaai25}

\newpage
\section{Appendix}
	\textbf{Summary.} In this section, we theoretically reveals that target error is influenced by source graph error, total variation distance, and model smoothness. In detail, we first establish a term to measure the robustness of a function $f$ under distribution shift constrained by the Wasserstein distance on the source distribution \( \mathbb{S} \) and target distribution \( \mathbb{T} \). Then, we can deduce that robustness of $f$ in each domain is bounded by its model smoothness. Finally, by applying the domain transfer theorem, we can get the transfer error is bounded by the variation distance, and model smoothness of each domain.
	\begin{proof}
		Consider the set \( B_W(\mathbb{S}, r) = \{S : W_\infty(\mathbb{S}, S) \leq r\} \), where \( W_\infty \) denotes the \(\infty\)-th Wasserstein distance~\cite{ruschendorf1985wasserstein}, a metric commonly used to quantify the difference between two probability distributions. The set \( B_W(\mathbb{S}, r) \) encompasses all distributions \( S \) that lie within a radius \( r \) of the reference distribution \( \mathbb{S} \), under the \( W_\infty \) metric.
		Next, define \( \mathbb{S}_r \) as the distribution within \( B_W(\mathbb{S}, r) \) that minimizes the expected loss \( \mathcal{E}_\mathbb{S}(f) \). Mathematically~\cite{li2022partial}, this can be expressed as:
		\begin{equation}
			\mathbb{S}_r = \arg\min_{S \in B_W(\mathbb{S}, r)} \mathcal{E}_\mathbb{S}(f).
		\end{equation}
		
		Similarly, define \( \mathbb{T}_r \) as the distribution within \( B_W(\mathbb{T}, r) \) that minimizes the expected loss \( \mathcal{E}_\mathbb{T}(f) \), given by:
		\begin{equation}
			\mathbb{T}_r = \arg\min_{T \in B_W(\mathbb{T}, r)} \mathcal{E}_\mathbb{T}(f).
		\end{equation}
		
		In essence, \( \mathbb{S}_r \) and \( \mathbb{T}_r \) denote the optimal distributions within their respective Wasserstein balls, \( B_W(\mathbb{S}, r) \) and \( B_W(\mathbb{T}, r) \), that yield the minimum expected loss for a given \( f \). 
		These distributions help assess the robustness of  \( f \) under distributional shifts constrained by the Wasserstein distance. Thus, we have the following derivation:
		\begin{equation}
			\begin{aligned}
				\mathcal{E}_{\mathbb{T}}(f) 
				&= \mathcal{E}_{\mathbb{T}}(f) - \mathcal{E}_{\mathbb{T}_r}(f) + \mathcal{E}_{\mathbb{T}_r}(f) 
				- \mathcal{E}_{\mathbb{S}}(f) + \mathcal{E}_{\mathbb{S}}(f)  \\
				&\leq \mathcal{E}_{\mathbb{S}}(f) + \Big| \mathcal{E}_{\mathbb{T}_r}(f) - \mathcal{E}_{\mathbb{T}}(f) \Big|
				+ \Big| \mathcal{E}_{\mathbb{T}}(f) - \mathcal{E}_{\mathbb{S}_r}(f) \Big|  \\
				&\leq \mathcal{E}_{\mathbb{S}}(f) + \Big| \mathcal{E}_{\mathbb{T}_r}(f) - \mathcal{E}_{\mathbb{T}}(f) \Big| 
				+ \Big|\mathcal{E}_{\mathbb{T}_{r}}(f) - \mathcal{E}_{\mathbb{S}_{r}}(f) \Big| \\
				&~~~~+ \Big|\mathcal{E}_{\mathbb{S}_r}(f) - \mathcal{E}_{\mathbb{S}}(f) \Big|.
			\end{aligned}
		\end{equation}
		
		Then, based on a recent study~\cite{yi2021improved}, for any model parameters $\boldsymbol{\theta}$  and $r$, we have the following:
		\begin{equation}
			\begin{aligned}
				\sup_{S \in B_{W}(\mathbb{S}, r)} R_S(\boldsymbol{\theta}) = \mathbb{E}_\mathbb{S} \left[ \sup_{\|\boldsymbol{\delta}\|_\infty \leq r} f(\boldsymbol{\theta}, \mathbf{x} + \boldsymbol{\delta}) \right].
			\end{aligned}
		\end{equation}
		
		This expression represents the supremum of the risk \( R_S(\boldsymbol{\theta}) \) over all distributions \( S \) within a Wasserstein ball \( B_{W_\infty}(\mathbb{S}, r) \), which equals the expected value under \( \mathbb{S} \) of the supremum of \( f(\boldsymbol{\theta}, \boldsymbol{x} + \boldsymbol{\delta}) \) for perturbations \( \boldsymbol{\delta} \) bounded by \( r\) in the infinity norm.
		It shows that the distributional perturbation measured by \( W_\infty \)  distance is equivalent to input perturbation. Hence, we can study \( W_\infty \) distributional robustness through \(\ell_\infty\)-input-robustness (i.e., model smoothness). 
		
		Inspired by~\cite{xu2012robustness}, we construct an \( r \)-cover for the metric space \( (\mathcal{X}, \| \cdot \|_2) \). The covering number \( \mathcal{N}(r, \mathcal{X}, \| \cdot \|_2) \) denotes the minimum number of \( r \)-balls needed to cover \( \mathcal{X} \) under the \( \ell_2 \)-norm. This satisfies the inequality \( \mathcal{N}(r, \mathcal{X}, \| \cdot \|_2) \leq (2d)^{(2\Gamma/r^2 + 1)} = N \), where \( \mathcal{X} \) can be enclosed by a polytope with \( \ell_2 \)-diameter smaller than \( 2\Gamma \) and \( 2d \) vertices. For details, refer to Theorem 4 in~\cite{vershynin2018high}.
		Next, we consider the space \( \mathcal{X} \) under the \( \ell_\infty \)-norm. Due to the geometric properties of \( \mathcal{X} \), the covering number under the \( \ell_\infty \)-norm satisfies a similar inequality \( \mathcal{N}(r, \dots) \).
				
		
		To construct such a cover explicitly, consider the collection of sets \( (C_1, \cdots, C_N) \), where each \( C_i \) is disjoint from the others and satisfies \( \| \textbf{u} - \textbf{v} \|_\infty \leq r \) for any \( \textbf{u}, \textbf{v} \in C_i \). The sets \( C_i \) can be constructed as \( C_i = \hat{C}_i \cap \left( \bigcup_{j=1}^{i-1} \hat{C}_j \right)^c \), where \( (\hat{C}_1, \cdots, \hat{C}_N) \) is a covering of \( \mathcal{X} \) under the \( \ell_\infty \)-norm, and the diameter of each \( \hat{C}_i \) is smaller than \( r \), ensuring that each \( C_i \) is a valid subset within the cover and remains disjoint from the others. Thus, we have:
		\begin{equation}
			\begin{aligned}
				&~~~~~\Big|\mathcal{E}_{\mathbb{S}_r}(f) - \mathcal{E}_{\mathbb{S}}(f) \Big| \\
				&=\left|  \sup_{S \in B_{W}(\mathbb{S}, r)} R_S(\boldsymbol{\theta}) - R_{S_m}(\boldsymbol{\theta}) \right| \\ 
				&=\left|  \mathbb{E}_\mathbb{S} \left[ \sup_{\| \boldsymbol{\delta} \|_\infty \leq r} f(\boldsymbol{\theta}, \mathbf{x} + \boldsymbol{\delta}) \right] - R_{S_m}(\boldsymbol{\theta})  \right|\\
				&= \left| \sum_{j=1}^N \mathbb{E}_\mathbb{S} \left[ \sup_{\| \boldsymbol{\delta} \|_\infty \leq r} f(\boldsymbol{\theta}, \mathbf{x} + \boldsymbol{\delta}) \bigg| \mathbf{x} \in C_j \right] \mathbb{S}(C_j) - R_{S_m}(\boldsymbol{\theta}) \right|\\
				&\leq \Bigg| \sum_{j=1}^N \mathbb{E}_\mathbb{S} \left[ \sup_{\| \boldsymbol{\delta} \|_\infty \leq r} f(\boldsymbol{\theta}, \mathbf{x} + \boldsymbol{\delta}) \bigg| \mathbf{x} \in C_j \right] \frac{|A_j|}{m} \quad \!\!\!\!\!\!  - \frac{1}{m} \sum_{i=1}^m f(\boldsymbol{\theta}, \mathbf{x}_i)  \\
				&\quad + \sum_{j=1}^N \mathbb{E}_\mathbb{S} \left[ \sup_{\| \boldsymbol{\delta} \|_\infty \leq r} f(\boldsymbol{\theta}, \mathbf{x} + \boldsymbol{\delta}) \bigg| \mathbf{x} \in C_j \right] \left( \frac{|A_j|}{m} - \mathbb{S}(C_j) \Bigg|   \right) \\
				&\leq \Bigg| \frac{1}{m} \sum_{j=1}^N \sum_{\mathbf{x}_i \in C_j} \sup_{\mathbf{x} \in C_j + B_\infty(0, r)} | f(\boldsymbol{\theta}, \mathbf{x}) - f(\boldsymbol{\theta}, \mathbf{x}_i) | \\
				&\quad + \Upsilon \sum_{j=1}^N \left| \frac{|A_j|}{n} - \mathbb{S}(C_j) \right| \Bigg| \\
				&\leq \frac{1}{m} \sum_{i=1}^n \sup_{\| \boldsymbol{\delta} \|_\infty \leq 2r} \Big| f(\boldsymbol{\theta}, \mathbf{x}_i + \boldsymbol{\delta}) - f(\boldsymbol{\theta}, \mathbf{x}_i) \Big| \\
				&\quad + \Upsilon \sum_{j=1}^N \left| \frac{|A_j|}{m} - \mathbb{S}(C_j) \right| \\
				&\leq \Phi_{\mathbb{S}} + \Upsilon \sum_{j=1}^N \left| \frac{|A_j|}{m} - \mathbb{S}(C_j) \right|.
			\end{aligned}
		\end{equation}
		
		Moreover, based on Proposition A6.6 in~\cite{van1996weak}, we have
		\begin{equation}
			\mathbb{P} \left( \sum_{j=1}^N \left| \frac{|A_j|}{m} - \mathbb{S}(C_j) \right| \geq \xi \right) \leq 2^N \exp \left( \frac{-m \xi^2}{2} \right).
		\end{equation}
		
		Combine this with Eq. (8), due to the value of $N$, we can draw the following conclusion:
		\begin{equation}
			\left|\mathcal{E}_{\mathbb{S}_r}(f) - \mathcal{E}_{\mathbb{S}}(f)\right| \leq \Phi_{\mathbb{S}}
			+ \Upsilon \sqrt{\frac{(2d)^{\frac{2\Phi^2 \Gamma}{r^2} + 1} \log 2 + 2 \log  (1/ \xi)}{m}}, 
		\end{equation}
		
		Similarly, due to the symmetry between the source distribution $\mathbb{S}$ and the target distribution $\mathbb{T}$, we can derive the following conclusion:
		\begin{equation}
			\left|\mathcal{E}_{\mathbb{T}_r}(f) - \mathcal{E}_{\mathbb{T}}(f)\right| \leq \Phi_{\mathbb{T}}
			+ \Upsilon \sqrt{\frac{(2d)^{\frac{2\Phi^2 \Gamma}{r^2} + 1} \log 2 + 2 \log  (1/ \xi)}{n}}, 
		\end{equation}
		
		Furthermore, based on the definition of TVD and Theorem 4 in~\cite{yi2021improved}, we have:
		\begin{equation}
			\left|\mathcal{E}_{\mathbb{T}_r}(f) - \mathcal{E}_{\mathbb{S}_r}(f)\right| \leq 2 \text{TVD}(\mathbb{S}, \mathbb{T}) 
			+ \Upsilon \sqrt{\frac{\log  (1/ \xi)}{2m}}.
		\end{equation}
		
		Finally, by summing Eq. (10), Eq. (11), and Eq. (12), we conclude with the following result:
		\begin{equation}
			\begin{aligned}
				\mathcal{E}_{\mathbb{T}}(f) & \leq \mathcal{E}_{\mathbb{S}}(f) + 2 \mathrm{TVD}(\mathbb{S}, \mathbb{T}) +  \Phi_{\mathbb{S}}+\Phi_{\mathbb{T}} \\
				&+ \Upsilon \sqrt{\frac{(2d)^{\frac{2\Phi^2 \Gamma}{r^2} + 1} \log 2 + 2 \log (1/ \xi)}{m}} \\
				&+ \Upsilon \sqrt{\frac{(2d)^{\frac{2\Phi^2 \Gamma}{r^2} + 1} \log 2 + 2 \log (1/ \xi)}{n}} \\
				&+ \Upsilon \sqrt{\frac{\log(1/ \xi)}{2m}}.
			\end{aligned}
		\end{equation}
		
		Therefore, we get our conclusion.
	\end{proof}

\end{document}